\documentclass[journal,comsoc]{IEEEtran}
\usepackage[T1]{fontenc}
\usepackage{lineno,hyperref}
\modulolinenumbers[5]
\usepackage{amsmath,amsfonts,amssymb}
\usepackage{verbatim}
\usepackage{algpseudocode}
\usepackage{graphicx}
\usepackage{textcomp} 
\usepackage{multirow} 
\usepackage{epstopdf}
\usepackage{color}
\usepackage{algorithm}
\usepackage{threeparttable}
\usepackage[table,xcdraw]{xcolor}
\usepackage{subfigure}
\usepackage{changepage}
\usepackage{bm}
\usepackage{diagbox}
\usepackage{float}

\newcommand{\fd}{\mathrm{d}}
\newcommand{\fr}{\mathrm{r}}

\newcommand{\bht}{\mathbf{h}}

\newcommand{\ut}{\mathbf{u}}
\newcommand{\wt}{\mathbf{w}}
\newcommand{\mTa}{\mathbf{\Theta}}

\newcommand{\Gt}{\mathbf{G}}
\newcommand{\Wt}{\mathbf{W}}
\newcommand{\Pt}{\mathbf{P}}
\newcommand{\Qt}{\mathbf{Q}}
\newcommand{\St}{\mathbf{S}}
\newcommand{\Tt}{\mathbf{T}}

\newcommand{\bbC}{\mathbb{C}}
\newcommand{\bbR}{\mathbb{R}}

\newcommand{\clN}{\mathcal{N}}

\begin{document}
\title{Hybrid Beamforming for RIS-Aided Communications: Fitness Landscape Analysis and Niching Genetic Algorithm}
\author{Bai~Yan,~
	Qi~Zhao,~	
	Jin~Zhang,~	
    J. Andrew~Zhang,~\IEEEmembership{Senior Member,~IEEE}~
	and~Xin~Yao,~\IEEEmembership{Fellow,~IEEE}	
\thanks{Corresponding author: Jin Zhang.}
\thanks{B. Yan and Q. Zhao are with Guangdong Provincial Key Laboratory of Brain-Inspired Intelligent Computation, Department of Computer Science and Engineering, Southern University of Science and Technology, Shenzhen 518055, China, and also with School of Computer Science and Technology, University of Science and Technology of China, Hefei 230027, China (email: yanb@sustech.edu.cn; zhaoq@sustech.edu.cn).}
\thanks{J. Zhang and X. Yao are with Guangdong Provincial Key Laboratory of Brain-Inspired Intelligent Computation, Department of Computer Science and Engineering, Southern University of Science and Technology, Shenzhen 518055, China (email: zhangj4@sustech.edu.cn; xiny@sustech.edu.cn).}
\thanks{J. A. Zhang is with Global Big Data Technologies Centre (GBDTC), University of Technology Sydney, NSW 2007, Australia (email: Andrew.Zhang@uts.edu.au).}}


\maketitle
\begin{abstract}
Reconfigurable Intelligent Surface (RIS) is a revolutionizing approach to provide cost-effective yet energy-efficient communications. The transmit beamforming of the base station (BS) and discrete phase shifts of the RIS are jointly optimized to provide high quality of service. However, existing works ignore the high dependence between the large number of phase shifts and estimate them separately, consequently, easily getting trapped into local optima. To investigate the number and distribution of local optima, we conduct a fitness landscape analysis on the sum rate maximization problems. Two landscape features, the fitness distribution correlation and autocorrelation, are employed to investigate the ruggedness of landscape. The investigation results indicate that the landscape exhibits a rugged, multi-modal structure, i.e., has many local peaks, particularly in the cases with large-scale RISs. To handle the multi-modal landscape structure, we propose a novel niching genetic algorithm to solve the sum rate maximization problem. Particularly, a niching technique, nearest-better clustering, is incorporated to partition the population into several neighborhood species, thereby locating multiple local optima and enhance the global search ability. We also present a minimum species size to further improve the convergence speed. Simulation results demonstrate that our method achieves significant capacity gains compared to existing algorithms, particularly in the cases with large-scale RISs.
\end{abstract}

\IEEEpeerreviewmaketitle

\section{Introduction}
\textit{Reconfigurable Intelligent Surface} (RIS), also known as intelligent reflection surface, is a planar artificial metasurface consisting of many passive reflecting elements \cite{9360709}. Each reflecting element independently induces an amplitude and/or phase shift to the incident signal, thereby collaboratively generating a directional beam to boost the link quality \cite{chen2016review}\cite{li2019machine}. Different from the conventional active relaying, RIS does not incur any noise amplification or require any active radio-frequency chains for signal transmission, hence RIS is more cost-effective and energy-efficient. As a result, the RIS has attracted increasing attention in many applications \cite{yuan2020intelligent}\cite{mishra2019channel}, such as wireless power transfer, cognitive radio network, wireless communication. 

Focusing on the potential of RIS techniques, some related works have been reported in point-to-point communications or multi-user systems. Most of these works assume that the RIS induces a certain phase shift by each element on the incident signal. Thus the RIS passive beamforming is equivalent to optimizing phase shifts. Optimization of the base station (BS) beamforming and RIS beamforming target at minimizing the transmit power of the BS \cite{wu2019intelligent}, maximizing the sum rate \cite{yu2019miso}, maximizing the weighted sum rate \cite{guo2020weighted}, or maximizing the energy efficiency \cite{huang2019reconfigurable}. The BS transmit beamforming is estimated based on the semidefinite relaxation technique \cite{wu2019intelligent} or zero forcing \cite{huang2019reconfigurable}. These works \cite{wu2019intelligent}-\cite{huang2019reconfigurable} assume continuous phase shifts, yet in practice only the discrete values are supported due to the hardware limitation. 

To achieve discrete phase shifts of RIS, a quantization approach is proposed in \cite{huang2018energy}\cite{chen2019intelligent} to achieve discrete value from the optimized continuous phase-shift setting. However, this approach often leads to unpredictable performance loss. Based on this approach, the work \cite{guo2019weighted} improves the performance by performing quantization at each iteration. In \cite{wu2019beamforming}\cite{di2020hybrid}, a sequential algorithm is proposed to optimize the element's phase shift one by one. A shortcoming of the sequential algorithm is that it easily gets trapped into uninteresting solutions. Alternatively, a branch-and-bound based method \cite{di2020hybrid} is developed and achieves performance improvements. Unfortunately, this method is not scalable to large-scale RISs as the scale of the computational complexity is $O(2^{bN})$, where $b$ and $N$ are the number of quantization bits and the number of RIS elements.  

For the sake of simplicity, the above works \cite{huang2018energy}-\cite{di2020hybrid} optimize the phase shifts separately. However, the discrete phase shifts essentially interact with each other and collaboratively contribute to the overall performance. Consequently, these works easily get trapped into local optima or even fail. It is highly desired to investigate the number and distribution of local optima in the landscape. 

\begin{figure}[t]
	\centering
	\includegraphics[width=7cm,height=4.5cm]{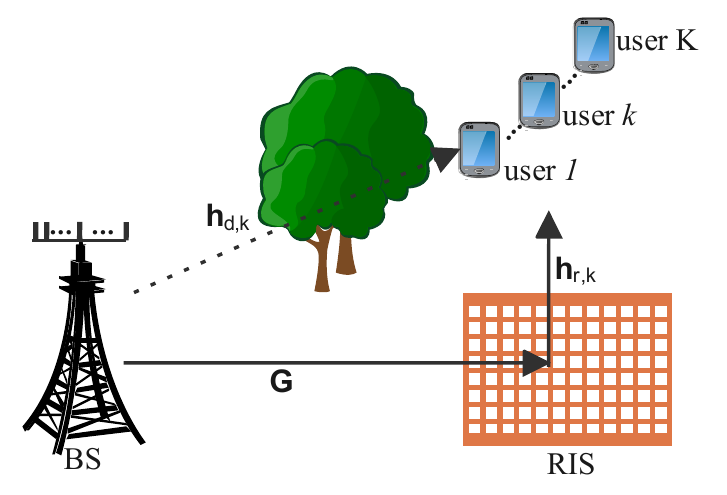}
	\caption{A downlink RIS-aided MU-MISO system.}
	\label{fig-system}
\end{figure}

Therefore, in this paper, we use fitness (objective function) landscape analysis techniques \cite{mersmann2011exploratory}\cite{malan2013survey} to investigate the influence of discrete phase shifts on the underlying optimization landscape. The analysis would provide guidance to better problem understanding and algorithm design. We consider a downlink RIS-aided multi-user multiple-input single-output (MU-MISO) system where the direct links between the BS and users suffer from deep shadowing, as shown in Fig. \ref{fig-system}. The transmit beamforming of the BS and the discrete RIS beamforming need to be jointly optimized to maximize the sum rate of users with limited transmit power. To analyze the landscape of this problem, we utilize two landscape features, including fitness distance correlation (FDC) and autocorrelation, to study the ruggedness of landscape from a global and local view respectively. It is shown that the underlying optimization problem gradually exhibits a rugged, multi-modal landscape structure as the RIS size increases, i.e., has many local peaks. The widespread local peaks would seriously degrade the performance of existing methods.

To tackle the NP-hard nature and multi-modal landscape structure, we propose a niching genetic algorithm (NGA) for sum rate maximization. We incorporate a well-known niching technique, the nearest-better clustering \cite{preuss2010niching}, into the genetic algorithm to handle the multi-modal property. The main idea is that the population is partitioned into several neighborhood species to locate multiple optima, hence falling into the local optima can be avoided and the global optima is well found. But if there are too many species, the whole convergence speed would slow down. To address this, we modify the nearest-better clustering by limiting the minimum size of each species. Our main contributions are summarized below:
\begin{itemize}
	\item We conduct fitness landscape analysis to investigate how the phase shifts affect the landscape of sum rate. Two landscape features, the FDC and autocorrelation are employed. Considering the periodicity of phase shifts, we propose three distance metrics for landscape analysis and find the \textit{cycle-1 distance} is the best. The landscape of sum rate exhibits a rugged, multi-modal structure as the RIS size increases. 
		
	\item Niching genetic algorithm. We incorporate the nearest-better clustering technique into NGA to partition the population into several neighborhood species, so as to maintain the solution space diversity and locate the global optima. A minimum species size is presented to better balance the convergence speed and global search capability.   
	
	\item Empirical validation of NGA's performance. Simulation results demonstrate that NGA achieves significant capacity gains compared to the state-of-the-art algorithms, particularly in the cases with a large-scale RIS.
	
\end{itemize} 
 
The rest of the paper includes the system model and problem formulation in Section II, and fitness landscape analysis for the sum rate maximization problems in Sections III. We propose a niching genetic algorithm in Section IV, and present simulation results in Section V. Section VI concludes the paper. 

\textit{Notation:} Low-case bold face variables and upper-case bold face ones represent vectors and matrices respectively. $(\cdot)^H$ and $(\cdot)^{-1}$ denote conjugate transpose and inverse, respectively. $\bbR$ and $\bbC$ are the real domain and complex domain.

\section{System Model and Problem Formulation}
\subsection{System Model}
We consider a downlink RIS-aided MU-MISO system where one multi-antenna BS serves $K$ single-antenna users (see Fig. \ref{fig-system}). The direct links between the BS and users suffer from serious fading or potential obstacles. Thus an RIS is deployed between the BS and users to provide high-quality links. We denote the baseband equivalent channels from the BS to user $k$, from BS to RIS, and from RIS to user $k$ as $\bht_{\fd,k}\in\bbC^{M\times1}$, $\Gt\in\bbC^{N\times M}$, and $\bht_{\fr,k}\in\bbC^{N\times1}$, where $M$ and $N$ are the numbers of BS antennas and RIS elements. In this paper, we only consider the discrete phase-shift setting, considering the hardware limitation. Assume the number of quantization bits for the phase shifts is $b$, each RIS element has $2^b$ possible modes. Then, the phase shift matrix is defined as a diagonal matrix $\mTa=diag(\theta_1,...,\theta_n,...,\theta_N)$ with 
\begin{equation}
	\theta_n=\beta_ne^{j\phi_n},\ \phi_n=\frac{\tau_n2\pi}{2^b}, \tau_n\in\{0,...,2^b-1\},
	\label{eq-theta-n}
\end{equation}
where $\beta_n$ and $\phi_n$ indicate the reflection amplitude and phase shift of element $n$. We call $\bm{\tau}=[\tau_1,...,\tau_n,...,\tau_N]$ as a ``\textit{configuration}''. For simplicity, we set $\beta_n$s to 1 \cite{wu2019beamforming}\cite{di2020hybrid}. We assume the channel state information (CSI) of all channels is perfectly known at the BS and RIS and the mobility of the users is very limited, which is the same as \cite{huang2018energy}-\cite{di2020hybrid}.

The received signal at user $k$ \cite{di2020hybrid} can be cast as 
\begin{equation}
	\label{eq-received-signal} 
	y_k=(\bht_{\fd,k}+\bht_{\fr,k}^H\mTa\Gt)\sum_{k=1}^{K}\wt_k s_k+\zeta_k,  	
\end{equation}
where $\wt_k\in\bbC^{m\times1}$ is the BS beamforming vector to user $k$. $s_k$ is the transmitted symbol to user $k$, an independent random variables with zero mean and unit variance. $\zeta_k$ denotes the additive white Gaussian noise at the user $k$ with zero mean and power $\sigma ^{2}$.

The $k$-th user treats all the signals from the remaining users as interference. Hence, the decoding signal-to-interference-plusnoise ratio (SINR) at the user $k$ can be represented as 
\begin{equation}
	\label{eq-SINRk}
	\begin{aligned}
		\gamma_k=\frac{|(\bht_{\fd,k}+\bht_{\fr,k}^H\mTa\Gt)\wt_k|^2}{\sum_{k\neq k'}| (\bht_{\fd,k}+\bht_{\fr,k}^H\mTa\Gt)\wt_k'|^{2}+\sigma ^{2}}. 
	\end{aligned}
\end{equation}

\subsection{Problem Formulation}
In this paper, we aim to maximize the sum rate of all users subject to the discrete phase shifts and the transmit power constraint:
\begin{subequations}
	\label{eq-maxSumRate}
	\begin{align}
        &\max_{\Wt, \mTa}\quad \sum_{k=1}^{K}\log _{2}(1+\gamma_k),\tag{\ref{eq-maxSumRate}{a}}\label{eq-Pa} \\ 
        &\quad s.t.\quad \theta_n=\beta_ne^{j\phi_n},\tag{\ref{eq-maxSumRate}{b}}\label{eq-Pb} \\ 
        &\qquad\quad\ \phi_n=\frac{\tau_n2\pi}{2^b}, \tau_n\in\{0,...,2^b-1\},\tag{\ref{eq-maxSumRate}{c}}\label{eq-Pc} \\
        &\qquad\quad\ \sum_{k=1}^{K}\|\mathbf{w}_k\|^{2}\leq P_{T},\tag{\ref{eq-maxSumRate}{d}}\label{eq-Pd}		
	\end{align}
\end{subequations}
where $\Wt=[\wt_1,...,\wt_K]\in\bbC^{M\times K}$ is the BS beamforming to all users, the equation (\ref{eq-Pd}) indicates that the transmit power is constrained below $P_T$. Problem (\ref{eq-maxSumRate}) is a mixed integer non-convex optimization problem, which is NP-hard. 

\section{Fitness Landscape Analysis for Problem (\ref{eq-maxSumRate})} \label{sec-fitness landscape analysis}
In this section, we perform a fitness landscape analysis to investigate how the phase shifts affect the landscape of sum rate. The analysis would help us better understand the optimization problem difficulties and provide insights for algorithm design.

\subsection{Fitness Landscape and Landscape Features}
Fitness landscape is a useful tool for understanding the structure of landscape, defining problem difficulties, and giving insights to algorithm design \cite{mersmann2011exploratory}\cite{malan2013survey}. It describes the relation between the search space and fitness space according to some randomly drawn problem instances. 

More formally, a discrete fitness landscape $(X,f,\clN_t)$ is composed of a set of solutions $x\in X$, a fitness function $f(x): X\rightarrow\bbR$, and a neighborhood $\clN_D(x)$ obtained over each solution by a distance metric $D$. For landscape analysis, the neighborhood and distance metric are required to define based on the specific problems. 

There are many features that define the structure of a fitness landscape. Here we focus on the landscape ruggedness. A fitness landscape is said to be rugged if the landscape contains many peaks, and if there is low correlation between neighboring points. It has been shown that the number of local optima increases with the ruggedness of a landscape \cite{merz2000fitness}. Here we employ the fitness distance correlation and autocorrelation to investigate the ruggedness of the landscape from a global and local view, respectively.   

\subsubsection{Fitness distance correlation (FDC)} FDC provides a global view of fitness landscapes and problem difficulty. It examines the correlation between fitness and the distance to the global/nearest optimum in the search space \cite{2005A}. Assume a random sample of solutions $\{x_l\}_{l=1}^L$, their fitness values are $F=\{f_l\}_{l=1}^L$ and the distances to the global optimum $x^*$ are $S=\{s_l\}_{l=1}^L$. The FDC is estimated by 
\begin{equation}
	\label{eq-FDC}
	\begin{aligned}
		\varrho(f,s)\approx\frac{1}{\sigma_f\sigma_s L}\sum_{l=1}^L(f_l-\bar{f})(s_l-\bar{s}),
	\end{aligned}
\end{equation}   
where $\bar{f}$ and $\bar{s}$ are mean values of $F$ and $S$, $\sigma_f$ and $\sigma_s$ represent the standard deviations of $F$ and $S$. For a maximization problem, if the fitness increases when the distance to the optimum becomes smaller, then the search is expected to be easy since there is a ``path'' to the optimum via solutions with increasing fitness. According to \cite{2005A}, if $\varrho\leq-0.15$, the fitness and the distance to the optimum are related, the problem is easy to catch the global optimum. If $-0.15<\varrho<0.15$, the search is difficult as there is no clear correlation between fitness and distance to the global optimum. If $\varrho\geq0.15$, the problem is misleading because as the fitness increases, the distance to the optimum increases too.

\subsubsection{Autocorrelation and correlation length} It measures the ruggedness of a landscape based on the degree of correlation between sampling points. A low correlation value indicates a rugged landscape, making the search for the optima difficult. We perform a random walk of length $J+1$ by starting at an arbitrarily chosen initial point and moving to a neighboring point at each step. Then a set of points  $\{x_j\}_{j=1}^J$ associated with fitness values $\{f_j\}_{j=0}^J$ are obtained. The autocorrelation is defined as the correlation of neighboring fitness values along a random walk:
\begin{equation}
	\label{eq-autocorrelation}
	\begin{aligned}
		\rho(s)=\approx\frac{1}{\sigma_f^2(J-\nu)}\sum_{j=1}^{J-\nu}(f(x_j)-\bar{f})(f(x_{j+\nu})-\bar{f}), \notag
	\end{aligned}
\end{equation}   
where $\nu$ is the step size between the two points. Based on autocorrelation, a correlation length $\dot{\rho}$ is further defined to directly represent the ruggedness of a fitness landscape. It is formulated as \cite{stadler1996landscapes}
\begin{equation}
	\label{eq-correlation length}
	\begin{aligned}
		\dot{\rho}=-\frac{1}{\ln(|\rho(1)|)},
	\end{aligned}
\end{equation}  
for $\rho(1)\neq0$. The lower $\dot{\rho}$, the rugged the landscape.

\subsection{Distance Metric and Neighborhood for Problem (\ref{eq-maxSumRate})} \label{sec-distance}
The fitness landscape relies on a distance metric and neighborhood function for analysis. Common distance choices are the number of local search moves between two solutions or the number of non-common variables in two solutions. However, these choices cannot be applied for problem (\ref{eq-maxSumRate}) due to the periodicity of phase shifts. From equation (\ref{eq-theta-n}), we know that the period of a phase shift $\phi_n$ is $2\pi$ and correspondingly, the period of each configuration element $\tau_n$ is $2^b$. Therefore, giving two configurations $\bm{\tau}_1$ and $\bm{\tau}_2$, we define a distance metric called ``\textit{cycle-$q$ distance}'': 
\begin{equation}
	\label{eq-distance}
	\begin{aligned}
		D_q(\bm{\tau}_1,\bm{\tau}_2)=\sum_{i=1}^{N} \left(\Omega([\bm{\tau}_1]_i,[\bm{\tau}_2]_i)\right)^q,
	\end{aligned}
\end{equation} 
where $[\bm{\tau}_1]_i$ is the $i$-th entry of $\bm{\tau}_1$, $\Omega(\cdot,\cdot)$ computes the minimum number of moves between the inputs considering the periodicity. The possible $q$ values are $\{0,1,2\}$, relating to the traditional Hamming distance, Cityblock distance and Euclidean distance, respectively. We will choose the best-fit distance for Problem (\ref{eq-maxSumRate}) in the next subsection. 

Take $\bm{\tau}_1=[0,1,1]$ and $\bm{\tau}_2=[3,1,3]$ with the number of quantization bits $b=2$ as an example. The cycle-$0$ distance is the number of non-common elements, i.e., $D_0(\bm{\tau}_1,\bm{\tau}_2)=2$. The cycle-$1$ distance is $D_1(\bm{\tau}_1,\bm{\tau}_2)=1+0+2=3$, where the minimum number of moves for each element ``0$\rightarrow$3'', ``1$\rightarrow$1'' and ``1$\rightarrow$3'' are ``1'', ``0'' and ``2'', respectively. Note that ``0$\rightarrow$3'' only needs 1 move rather than 3 moves due to periodicity. We can also obtain the cycle-$2$ distance is $D_2(\bm{\tau}_1,\bm{\tau}_2)=1^2+0^2+2^2=5$.

We also need to define a neighborhood for fitness landscape analysis. As a very few quantization bits ($b=1,2,3$) are usually considered for the RIS in practice, we define two configurations $\bm{\tau}_1$ and $\bm{\tau}_2$ as neighbors if $D_1(\bm{\tau}_1,\bm{\tau}_2)=1$. 
	
\begin{table}[t]
	\caption{Experimental settings for Problem (\ref{eq-maxSumRate}).}
	\label{tab-experiment-setting}
	\renewcommand{\arraystretch}{1.5}
	\centering
	\begin{tabular}{lllll}
		\hline
		Parameters & Values \\ \hline	
		BS location   & (0m, 0m)  \\
		RIS location  & (100m, 0m)\\
		Path-loss for BS-RIS and RIS-user link    & 20+20lg$d$ \\
		Path-loss for BS-user link   & 32.6+36.7lg$d$ \\
		Transmission bandwidth & 180kHz \\   
		Noise power spectral density & -170dBm/Hz \\
		SNR (defined as $P_T/\sigma^2$) & 2dB \\ \hline
	\end{tabular}
\end{table}

\subsection{Landscape Analysis for Problem (\ref{eq-maxSumRate})} \label{sec-landscape anylysis for sum rate}
We consider the downlink RIS-aided MU-MISO sytem illustrated in Fig. \ref{fig-system}, in which 4 antennas, 4 users uniformly and randomly located in a circle centered at $(100m,30m)$ with radius $10m$. The number of quantization bits is $b=2$ if not stated. Other system parameters are exhibited in Table \ref{tab-experiment-setting}. 

To perform a fitness landscape analysis on Problem (\ref{eq-maxSumRate}), we need to generate a number of random solutions to calculate the landscape features. Each solution is composed by a RIS configuration and a transmit beamforming of the BS. We focus on investigating the effect of RIS configurations on the sum rate's landscape regardless of the transmit beamforming of the BS. To do this, each RIS configurations is generated at random, then we employ the zero forcing together with power allocation \cite{hochwald2005vector} \footnote{We consider zero forcing because it has very low complexity, which is very suitable for large-scale antenna systems. Other techniques for digital beamforming such as MMSE can also be employed here.} to obtain a near optimal BS beamforming for this RIS configuration. Specifically, according to the zero forcing method, the transmit beamforming for the BS is obtained by
\textbf{\begin{equation}
		\label{eq-zero forcing-F}
		\begin{aligned}
			\Wt=\mathbf{F}^H(\mathbf{F}\mathbf{F}^H)^{-1}\Pt^{\frac{1}{2}}=\hat{\Wt}\Pt^{\frac{1}{2}},
		\end{aligned}
\end{equation}}
where $\mathbf{F}=[\mathbf{f}_1,..., \mathbf{f}_k, ..., \mathbf{f}_K]$ is the transmission matrix with $\mathbf{f}_k=\bht_{\fd,k}+\bht_{\fr,k}^H\mTa\Gt$; $\hat{\Wt}=\mathbf{F}^H(\mathbf{F}\mathbf{F}^H)^{-1}$. $\Pt$ is a diagonal matrix whose $k$-th diagonal element is the allocated power for the $k$ users, i.e., $p_k$. The zero forcing requires to satisfy two constraints: $|\mathbf{f}_k^H\wt_k|=\sqrt{(p_k)}$ and $|\mathbf{f}_k^H\wt_k'|=0$, $k\neq k'$. The BS beamforming optimization problem is reduced to a power allocation problem \cite{tse2005fundamentals}
\textbf{\begin{equation}
		\label{eq-zero forcing-water-filling}
		\begin{aligned}
			\max_{\{p_k\geq 0\}} &\sum_{1\leqslant k\leqslant K} \log_{2}(1+\frac{p_k}{\sigma^2}), \\
			s.t.\ &Tr(\Pt^{\frac{1}{2}}\hat{\Wt}^H\hat{\Wt}\Pt^{\frac{1}{2}})\leqslant P_T,
		\end{aligned}
\end{equation}}
where $Tr(\cdot)$ calculates the trace. The optimal solution to this problem can be acquired by water-filling as $p_k=\frac{1}{v_k}\max\{\frac{1}{\delta}-v_k\sigma^2,0\}$, where $v_k$ is the $k$-th diagonal element of $\hat{\Wt}^H\hat{\Wt}$ and $\delta$ is a normalized factor satisfying $\sum_{1\leqslant k\leqslant K} \max\{\frac{1}{\delta}-v_k\sigma^2,0\}=P_T$. 

To calculate the fitness landscape feature FDC, we consider $10^5$ random samples of RIS configurations. The FDC requires that the global optima are known (or a very near-global optimum). However, the global optima is very hard to obtain, since we need to enumerate all the possible RIS configurations, which is too time-consuming. Here we run the genetic algorithm 100 times with a population size 200 and maximum iterative generations $10^4$, and select the solution with the best performance as the near-global optimum. 

\begin{figure}[t] 
	\centering
	\subfigbottomskip=0pt
	\subfigure[N=20]{\includegraphics[width=0.52\textwidth]{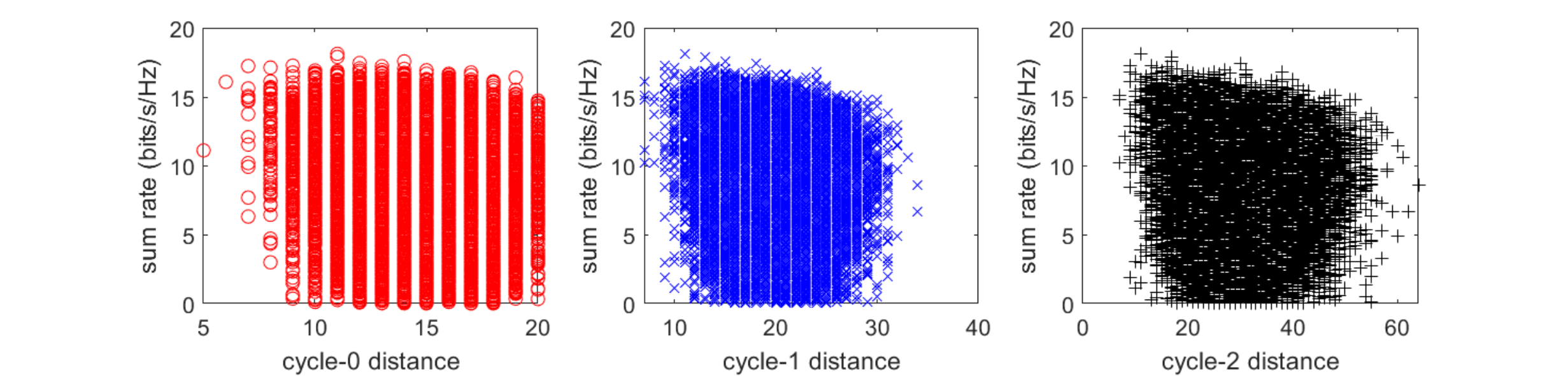}} \\ 
	\subfigure[N=80]{\includegraphics[width=0.52\textwidth]{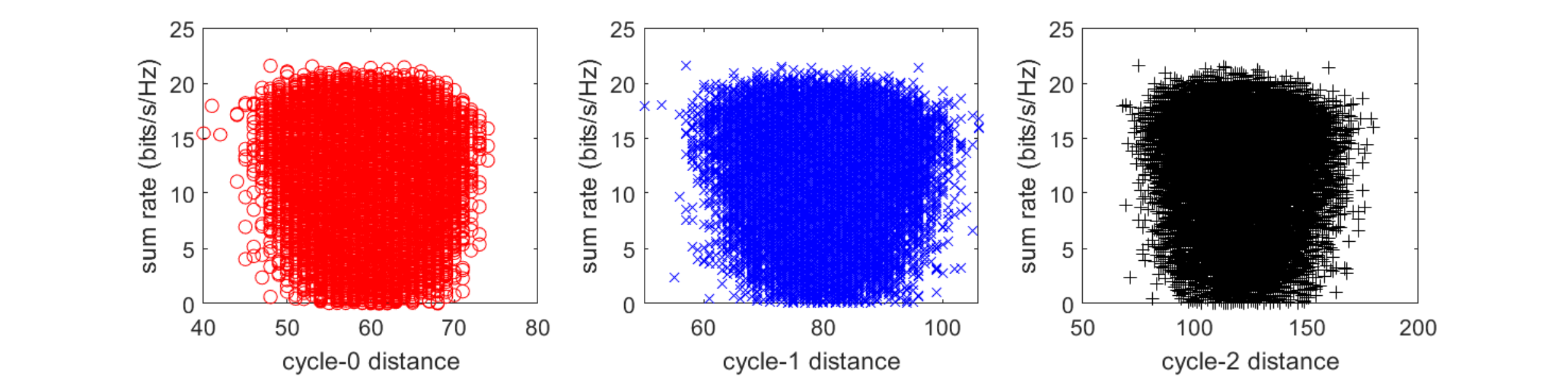}}\\
	\subfigure[N=140]{\includegraphics[width=0.52\textwidth]{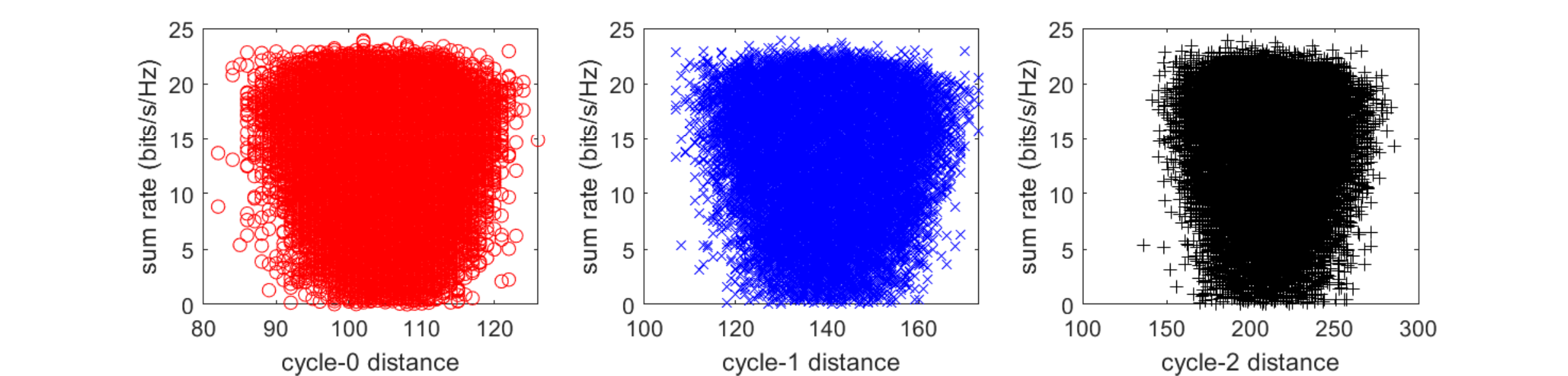}} \\
	\subfigure[N=200]{\includegraphics[width=0.52\textwidth]{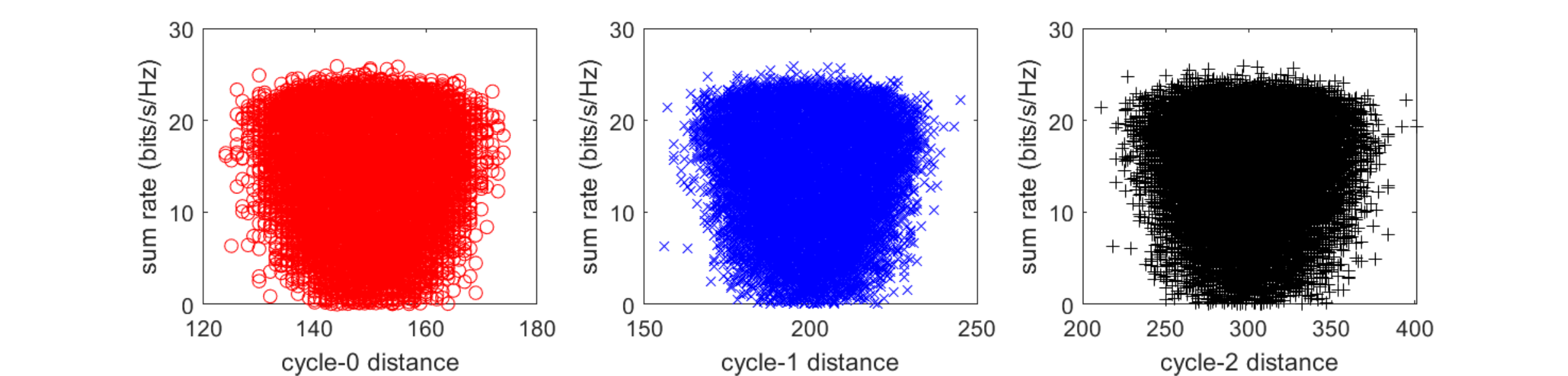}} \\
	\caption{FDC plots under three cylce-$q$ distance metrics with varying numbers of RIS elements $N$.} 
	\label{fig-FDC}
\end{figure}

Fig. \ref{fig-FDC} and Table \ref{tab-FDCs} shows the FDC plots and FDC values under three cylce-$q$ distance metrics with varying numbers of RIS elements. The FDC plots with $N=20$ show a weak correlation since the fitness increases as the distance to the nearest optimum is smaller. As $N$ is larger, the FDC plots exhibit a scatter point distribution and the corresponding FDC values are close to zero, so there is no correlation between the optimum values and the distance to the optimum solution. In addition, the FDC values increase as the number of phase shifts grows, indicating a more difficult optimization problem. Among the three distance metrics, cycle-1 distance achieves the highest correlation coefficients. 

For computing the autocorrelation, we consider a number of $10^5$ random walks and set the step size of each walk is 200. Table \ref{tab-correlation-entropys} lists the correlation lengths $\dot{\rho}$s and the normalized correlation lengths in relation to the diameter of the landscape $\frac{\dot{\rho}}{N}$s under different numbers of RIS elements $N$ and different number of quantization bits $b$. Here we focus more on $\frac{N}{\dot{\rho}}$, as it describes the ruggedness more fairly and accurately by removing the effect of the search space's diameter, compared to the correlation lengths $\dot{\rho}$. The shorter correlation length $\dot{\rho}$, the farther the value $\frac{\dot{\rho}}{N}$ to 0, the rugged landscape, the more local peaks. From table \ref{tab-correlation-entropys}, we can observe that, for each $b$, the normalized correlation length gets shorter as the number of RIS elements $N$ grows, indicating a more rugged landscape. For each $N$, as the number of quantization bits $b$ is larger, the normalized correlation length increases too, since the phase shift change between neighboring modes decreases.


\begin{table}[]
	\caption{FDC values under three cylce-$q$ distance metrics with varying numbers of RIS elements $N$.}
	\label{tab-FDCs}
	\renewcommand{\arraystretch}{1.5}
	\centering
	\begin{tabular}{llll}
		\hline
		\multicolumn{1}{c}{\multirow{2}{*}{$N$}} & \multicolumn{3}{c}{FDC} \\ 
		\multicolumn{1}{c}{}  & cycle-0 distance    & cycle-1 distance    & cycle-2 distance    \\ \hline
20       &  -0.0856 &  -0.1352  &-0.1180    \\
80     & -0.0269 & -0.0559 & -0.0437      \\
140    & -0.0175 & -0.0425 &-0.0300   \\
200   &  -0.0125 &  -0.0166 &  -0.0163   \\   \hline
	\end{tabular}
\end{table}

\begin{table}[]
	\caption{Correlation lengths $\dot{\rho}$ and normalized correlation lengths $\frac{\dot{\rho}}{N}$ under different numbers of RIS elements $N$ and different number of quantization bits $b$.}
	\label{tab-correlation-entropys}
	\renewcommand{\arraystretch}{1.5}
	\centering
	\begin{tabular}{lllllll}
		\hline
		\multicolumn{1}{c}{\multirow{2}{*}{$N$}} & \multicolumn{2}{c}{$b=1$} & \multicolumn{2}{c}{$b=2$}& \multicolumn{2}{c}{$b=3$}\\
	    \multicolumn{1}{c}{}  & \quad $\dot{\rho}$ &  \quad $\frac{\dot{\rho}}{N}$   & \quad $\dot{\rho}$ &  \quad \textbf{$\frac{\dot{\rho}}{N}$}  & \quad  $\dot{\rho}$ & \quad $\frac{\dot{\rho}}{N}$    \\ \hline
		20     & 1.8159 & 0.0908  & 3.1102 & 0.1555 & 10.1738  &  0.5087      \\
		80    & 3.6636 & 0.0458 & 6.7771 & 0.0847  & 20.9662  & 0.2621    \\
		140   & 5.3704 & 0.0384 & 10.3322 & 0.0738 & 31.5773  &  0.2256  \\
		200  & 6.8958 &  0.0345  & 12.9947 &  0.0650 & 39.7879 & 0.1989\\   \hline
	\end{tabular}
\end{table}

\subsection{Discussions}
The FDC results imply that there is no correlation between fitness and the distance to the global optimum. The correlation length results demonstrate a more rugged landscape when the RIS size is larger. It means that the Problem (\ref{eq-maxSumRate}) has a multi-modal landscape, i.e., there exists multiple local optimal solutions. As a result, existing optimization methods easily get trapped into local optima, achieving suboptimal or even unsatisfying performance. Moreover, decoupling the phase shifts ignores the high dependence between variables, further deteriorating the performance. Therefore, a more powerful algorithm with strong global search ability is highly desired to tackle the multi-modal property and non-separable variables. 

\section{Niching Genetic Algorithm for Problem (\ref{eq-maxSumRate})}
\subsection{Motivation}

Evolutionary algorithms are population-based optimizers with meta-heuristic search nature. They show great superiority in solving the NP-hard, non-convex, and non-separable problems. However, standard evolutionary algorithms are not suitable for Problem (\ref{eq-maxSumRate}) due to its weak ability of handling the multi-modal property. Thanksfully, the well-known niching techiques \cite{li2016seeking}\cite{das2011real} have been developed to modify the behavior of evolutionary algorithms so as to locate and preserve multiple optimal solutions. The niching techiniques include crowding, fitness sharing, restricted trounament selection, speciation, and clustering. The main idea of these techniques is to partition the whole population into several subpopulations to locate multiple optima. The global optima is then found from these optima. Therefore, combining the advantages of evolutionary algorithms and niching techniques, we propose a niching genetic algorithm (NGA) to solve Problem (\ref{eq-maxSumRate}).

\subsection{Overall Framework}
The overall process of NGA is outlined in Algorithm \ref{al-NGA}. It starts with initialization. A number of $N_Q$ phase shifts $\bm{\tau}$s for RIS are randomly generated. The BS beamforming $\Wt$s with respect to the phase shifts $\bm{\tau}$s are estimated via the zero forcing method (i.e., equations (\ref{eq-zero forcing-F}) and (\ref{eq-zero forcing-water-filling}). $\bm{\tau}$s and $\Wt$s constitute the initializtion population $\Qt^0$. The sum rate of $\Qt^0$ is computed according to equation (\ref{eq-Pa}. Iterative generations follow the initialization. In each generation, the nearest-better clustering technique is employed to partition the whole population into several neighborhood species. A minimum size $N_{min}$ is set to limit the scale of each species to provide a faster convergence speed to the global optima.
Thereafter, for each solution in a species, we employ the uniform crossover \cite{semenkin2012self} and random-resetting mutation \cite{kramer2017genetic} to produce a new solution. If the new solution achieves higher sum rate compared to the old one, the new solution survives and is stored into the next-generation population $\Qt^{t+1}$; otherwise, the old solution is put into $\Qt^{t+1}$. Once the iterative generation terminates, the NGA returns the solution with the highest sum rate. 

The core components, including the nearest-better clustering and solution reproduction, are detailed below.

\begin{algorithm}[t]
	\caption{The overall process of NGA} 
	\begin{algorithmic}[1]
		\Require $\bht_{\fd,k}^H$, $\bht_{\fr,k}^H$, $\Gt$, population size $N_Q$; 
		\Ensure	$\Wt$, $\mTa$;
		\State $t=0$;
		\State $\Qt^t\leftarrow$\textit{Initialization}$(N_Q)$;	
		\While {``\textit{the stopping criterion is not met}''}			
		\State $\St\leftarrow$\textit{Species\_Partition}($\Qt^t$); // Algorithm \ref{al-NBC-species-control}
		\For{each species in $\St$} 		
		\For{each configuration $\bm{\tau}_j$ in the current species}	
		\State $\ut_j\leftarrow$\textit{Reproduction}($\bm{\tau}_j$); // Section (\ref{sec-reproduction})
		\State select the solution with higher sum rate from 
		\Statex \quad\quad\quad\quad\quad $\{\bm{\tau}_j,\ut_j\}$ and put it into $\Qt^{G+1}$; 
		\EndFor
		\EndFor
		\State $t=t+1$;	
		\EndWhile	
		\State $\Wt, \mTa\leftarrow$find the solution with highest sum rate from $\Qt^t$.			
	\end{algorithmic}
	\label{al-NGA}
\end{algorithm}

\subsection{Nearest-Better Clustering with Minimum Species Size}
Nearest-better clustering \cite{preuss2010niching} is a commonly technique for multi-modal optimization problems. It partitions the whole population into several neighborhood species, and each species try to locate a different optima. 

After population partition, some species may only contain a few solutions, e.g., one or two. The crossover operator is unable to be executed in this case and the convergence speed would be slow. Conversely, if the species contain too may solutions, the population converge only a few local optima and may miss the true global optima. To well balance the convergence speed and performance, we carefully control the minimum size of each species, denoted as ``$N_{min}$''. A small value should be taken as $N_{min}$ at early iterations and a larger value at later iterations. The reason is: at early iterations, the population has little knowledge about the location of peaks, more species are required; as the iteration proceeds, a larger $N_{min}$ allows the species that locate the local optima to be combined into the neighborhood species that locate the global optima. Thus, the species converge to the global optima more quickly. In this paper, we set $N_{min}$ as
\begin{equation}
	\label{eq-species-size}
	\begin{aligned}
		N_{min}=5+\frac{t}{t_{max}}\times5,
	\end{aligned}
\end{equation} 
where $t$ and $t_{max}$ are the current and maximum generation. 

The process of NBC with minimum species size is shown in Algorithm \ref{al-NBC-species-control}. It contains the following three steps.

i) Preparation (Lines 1-3). The minimum species size is set by equation (\ref{eq-species-size}). The solutions are re-ordered by sum rate in descending order, and their mutual distance are calculated by the proposed cycle-1 distance. 

ii) Spanning tree construction (Lines 4-7). Except for the best solution, each solution is linked to its own nearest-better neighbor by creating an ``edge''. All the edges construct a spanning tree $\Tt$. For convenience, we call the nearest-better solution and the solution itself at each edge as ``leader'' and ''follower'', respectively. 

iii) Edges Cutting-off (Lines 8-23). The mean distance of all edges $\mu$ is calculated. All the edges are sorted by their distance in descending order. To find which edge to cut off, a \textit{follow} vector is defined. Each element of \textit{follow}($\cdot$) counts the number of solutions in the subtree rooted at the corresponding solution; each value of \textit{follow} is initialized to 1. For each edge, the \textit{follow} value of the leader solution is the sum of that of each follower solution. By analyzing the \textit{follow} before and after cutting off, the edges would be cut off to partition the population if satisfying the conditions: the distance of an edge is larger than the weighted mean distance $\varphi\times\mu$ and the species size after cutting off is no less than $N_{min}$. $\varphi$ is the weight on $\mu$, which controls the number of species. Larger $\varphi$ leads to fewer species. In this paper, we set $\varphi=1$.

\begin{algorithm}[t]
	\caption{Nearest-better Clustering with species size limit} 
	\begin{algorithmic}[1]
		\State Set $N_{min}$ by equation (\ref{eq-species-size});
		\State Sort the solutions from $\Qt$ by sum rate from the highest to the lowest;	
		\State Compute the mutual distances of all RIS configurations from $\Qt$ by the cycle-1 distance in equation (\ref{eq-distance}); 		
		\State Create an empty tree $\Tt$;
		\For {each RIS configuation $\bm{\tau}_i\in\Qt$}
		\State Find the nearest better neighbor $\bm{\tau}_{i,nb}$, build an edge between $\bm{\tau}_i$ and $\bm{\tau}_{i,nb}$, and add this edge into $\Tt$;
		\EndFor	
		\State Compute the mean distance $\mu$ of all edges in $\Tt$;
		\State Sort all edges by the cycle-1 distance in descending order;
		\For{each edge $e\in\Tt$}		
		\If{the distance of the edge > $\varphi\times\mu$}
		\State Collect the follower solutions of $e$ as $e_f$;
		\State Set the root of the subtree containing $e_f$ as $e_r$;
		\If{ $\textit{follow}(e_f)\geq N_{min}$ \textbf{and} \\ \qquad\quad\quad\ $\textit{follow}(e_r)-\textit{follow}(e_f)\geq N_{min}$} 
		\State Cut off $e$;
	    \State Set the leader solution of $e$ as $e_l$;
	    \For{each solution $s$ between $e_l$ and $e_f$}
	    \State \textit{follow}($s$)=\textit{follow}($s$)-\textit{follow}($e_f$);
	    \EndFor
		\EndIf
		\EndIf
		\EndFor
	\end{algorithmic}
	\label{al-NBC-species-control}
\end{algorithm}

\subsection{Solution Reproduction}\label{sec-reproduction}
We employ the \cite{semenkin2012self} and random-resetting mutation \cite{kramer2017genetic} to produce new solutions. The details are as follows.

\begin{figure}[t]
	\centering
	\includegraphics[width=8.6cm,height=1.8cm]{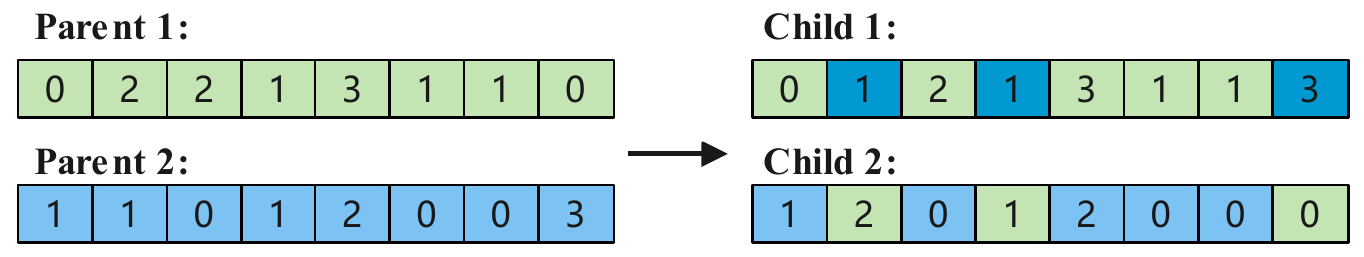}
	\caption{Uniform crossover. Assume $p_{cr}=0.7$ is used to decide exchange, and the array $[0.1, 0.7, 0.4, 0.9, 0.2, 0.1, 0.1, 0.8]$ is randomly generated.}
	\label{fig-crossover}
\end{figure}

For a candidate configuration $\bm{\tau}_j$, a crossover mate $\bm{\tau}_{j,mate}$ is randomly selected from the species that $\bm{\tau}_j$ belongs to. The uniform crossover works by treating each position of the configuration independently and choosing to exchange the entry with a probability $p_{cr}$. Take Fig. \ref{fig-crossover} as an example. $\bm{\tau}_j$ and $\bm{\tau}_{j,mate}$ correspond to Parent 1 and 2 in the figure. A array of random numbers with length $N$ is generated, e.g., $[0.1, 0.7, 0.4, 0.9, 0.2, 0.1, 0.1, 0.8]$. $p_{cr}$ is set as 0.7. Then, the parents exchange their entries where the random number is not less than 0.7. Therefore, the 2nd, 4th and last entry of the parents are exchanged, producing two children solutions.   

After crossover, the random-resetting mutation is performed on each children solution. Each position of a child configuration is replaced by other possible values with a probability $p_{mu}$. Here is an example of the random-resetting mutation:
\begin{equation}
	\label{eq-mutation}
	\begin{aligned}\nonumber
	\ [0, 1, \boxed{2}, 1, 3, 1, 1, \boxed{3}]\rightarrow[0, 1, \boxed{0}, 1, 3, 1, 1, \boxed{1}].
	\end{aligned}
\end{equation}

\subsection{Complexity Analysis}
The main computational complexity of NGA lies in the zero forcing and population partition. In the zero forcing method, optimizing the received power for each user requires $O(N_QK)$ computations, where $N_Q$ and $K$ are the population size and the number of users. In population partition, there are two time-consuming operations. One is calculating the cycle-1 distances between any two solutions, the other is to check whether the edge should be cut off to satisfy minimum species size. The time complexity of the two operations is $O(N_Q^2\times N)$ and $O(N_Q)$, where $N$ is the number of RIS elements. Thus, the total complexity is $O(N_Q^2\times N)$, which is much lower than the Branch-and-bound-based method \cite{di2020hybrid}.

\section{Experimental Study}
In this section, we conduct simulation experiments to validate the effectiveness of the proposed NGA for the RIS-aided MU-MISO system (as shown in Fig. \ref{fig-system}). The system parameters are set the same as that in Section \ref{sec-landscape anylysis for sum rate} if not stated. We compare the NGA to the following algorithms under various scenarios. All the comparison algorithms employ the zero forcing method (equations (\ref{eq-zero forcing-F}) and (\ref{eq-zero forcing-water-filling})) to estimate the transmit beamforming at the BS, as same as the NGA does. They differ in optimizing the discrete phase shifts:

\begin{itemize}
	\item Without RIS. In this case, there is no need to optimize the phase shifts.
	
	\item Sequential search method \cite{di2020hybrid}. At each iteration, keeping the other $N-1$ phase shifts fixed, the optimal value for the remaining phase shift is found by traversing all possible values. Repeat this until all the phase shifts are obtained.  
	
	\item Simulated annealing method \cite{di2020hybrid}\cite{di2016sub}. This method is a random-search heuristic method. We utilize the random-setting mutation to yield a new solution. The new solution is allowed to accept with a probability even if it does not bring performance enhancement.   
	
	\item Genetic algorithm. It is a population-based heuristic method that belongs to the evolutionary algorithms. Same as the NGA, the uniform crossover and random-resetting mutation are employed to produce new solutions.	
\end{itemize} 

For the genetic algorithm and NGA, we set the population size $N_Q=40$, crossover probability $p_{cr}=0.7$, and mutation probability $p_{mu}=0.01$. For a fair comparison, all the algorithms stop running if the maximum number of sum rate evaluations exceed 40000 or the change of sum rate is less than $10^{-6}$ in five consecutive iterations. The total number of Monte Carlo runs are set to 100 for all the algorithms.

\subsection{Investigation of Key Operators of xxx}

\begin{figure}[t]
	\centering
	\includegraphics[width=7cm,height=5.5cm]{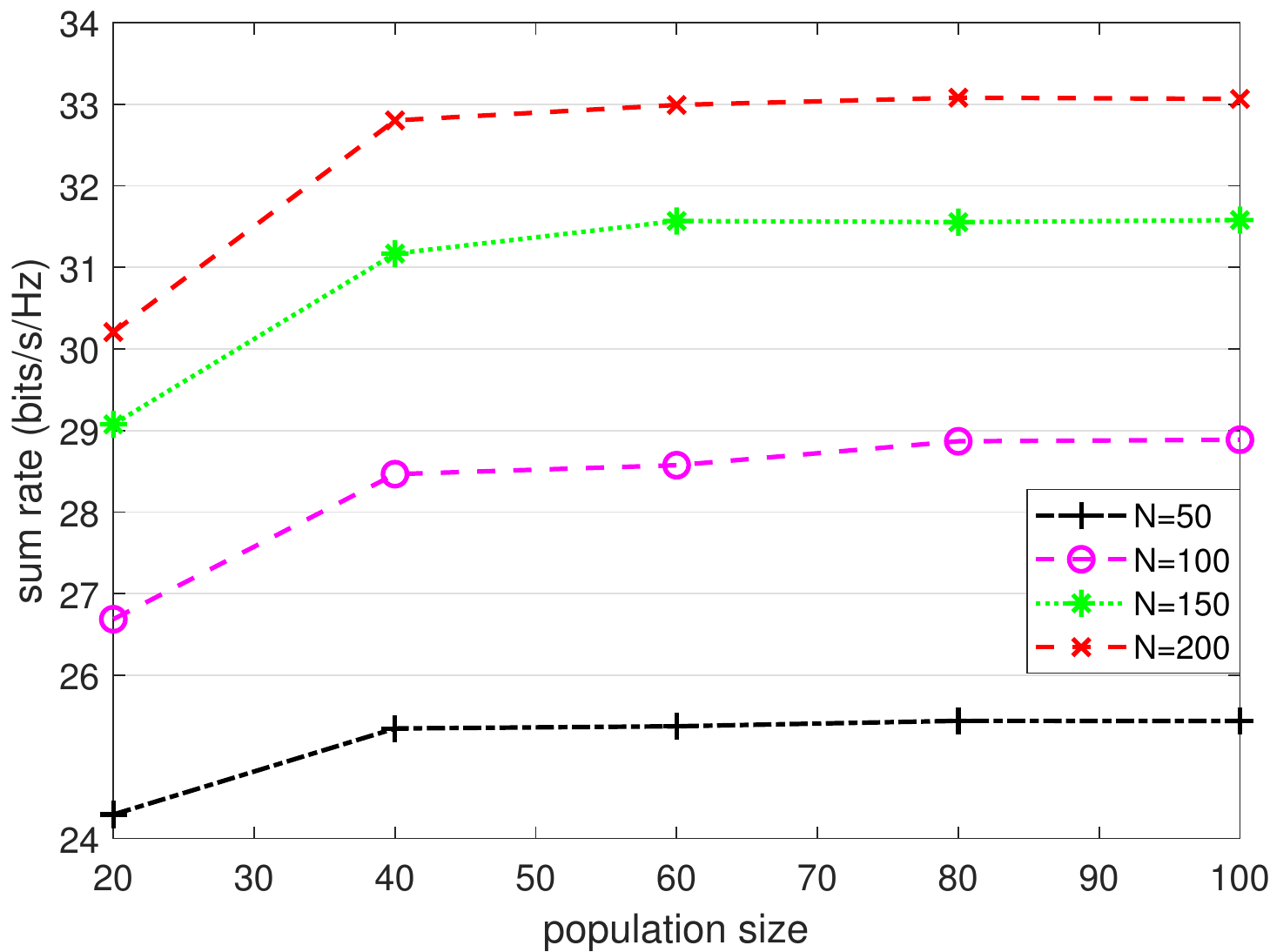}
	\caption{The sum rate versus the population size $N_Q$ with SNR=2dB and $b=2$.}
	\label{fig-reNP}
\end{figure}

\textbf{Parametric sensitivity}. Fig. \ref{fig-reNP} investigates the performance of NGA to the sensitivity of population size $N_Q$ with SNR=2dB and $b=2$. From this figure, we can observe that $N_Q$ has no influence on the sum rate when the number of RIS elements $N=50$. When $N$ belongs to $\{100, 150, 200\}$, the NGA achieves about 1dB gain as the population size increases from 20 to 40, then flattens as the population size continues increasing. Therefore, considering the balance between the performance and time complexity, we choose $N_Q=40$ for our method.

\begin{figure}[t]
	\centering
	\includegraphics[width=7cm,height=5.5cm]{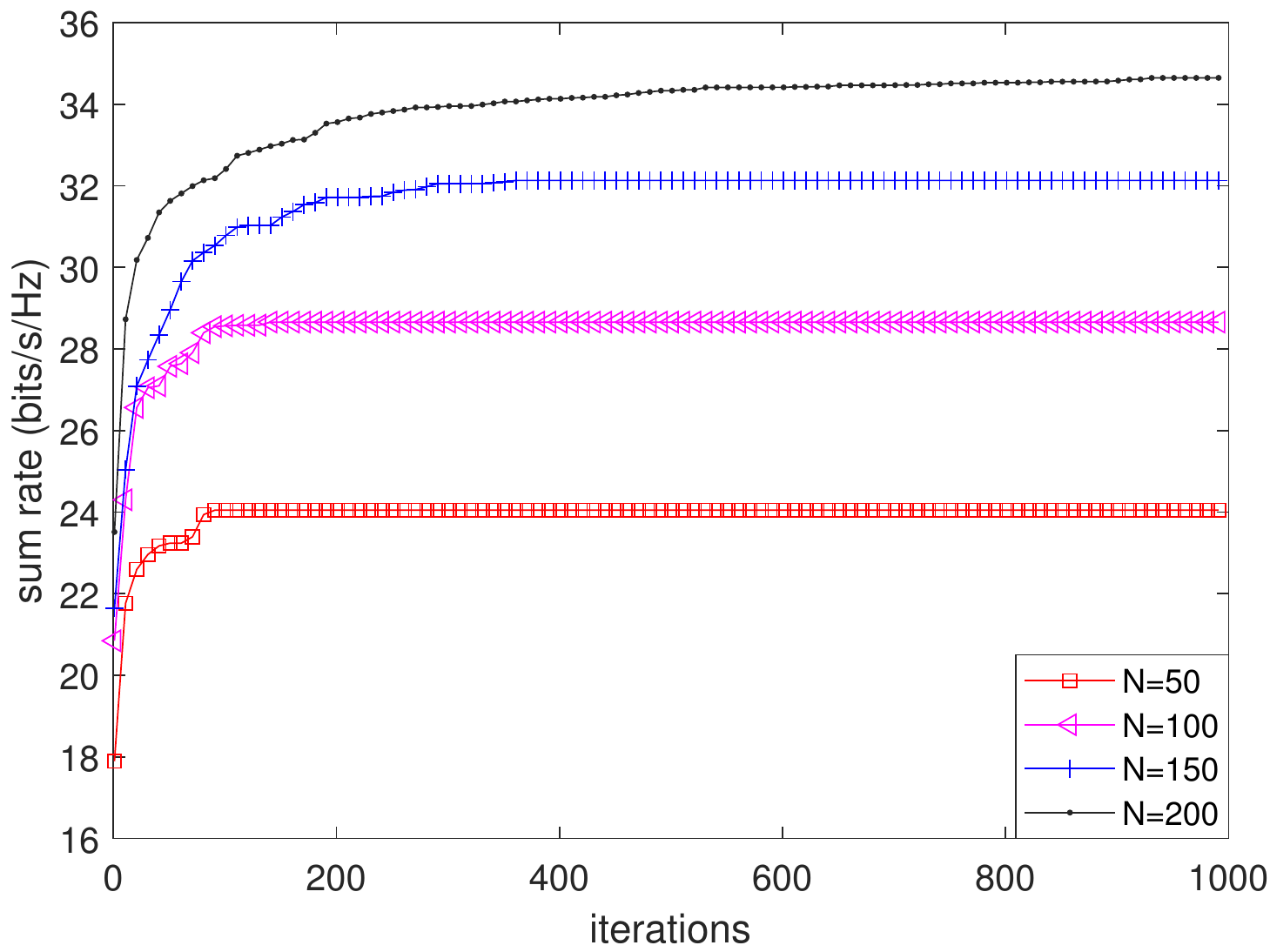}
	\caption{The typical convergence behavior under various numbers of RIS elements $N$ with SNR=2dB and $b=2$.}
	\label{fig-convergence-curve}
\end{figure}

\textbf{Convergence behavior}. Fig. \ref{fig-convergence-curve} shows the typical convergence behavior of NGA for different number of RIS elements $N$ with SNR=2dB and $b=2$. We observe that when the $N$ value takes 50, 100, 150 and 200, the NGA requires about 100 to 500 iterations. The fewer RIS elements, the faster convergence speed. Considering the parallel nature of evolutionary algorithms \cite{2015Distributed}, we suggest to accelerate NGA by parallel implementation to satisfy real-world applications. 

\subsection{Comparison of Our Method Against Other Algorithms}
\begin{figure}[t]
	\centering
	\includegraphics[width=7cm,height=5.5cm]{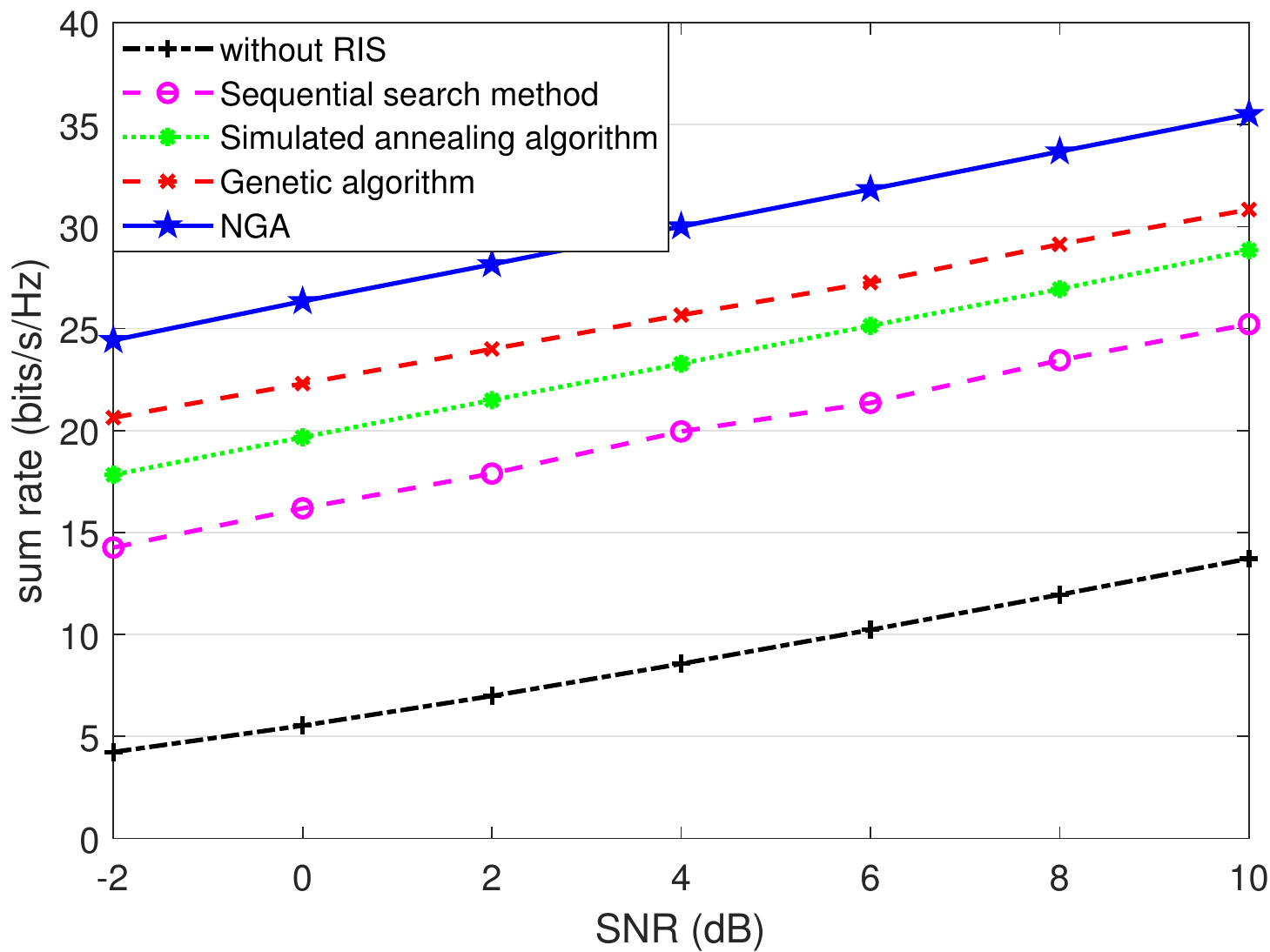}
	\caption{The sum rate versus SNR with $N=100$ and $b=2$.}
	\label{fig-reSNR}
\end{figure}

\textbf{Effect of SNRs}. Fig. \ref{fig-reSNR} shows the sum rate versus SNR with the number of RIS elements $N=100$ and the number of quantization bits $b=2$. It can be observed that the sum rate increases with SNR since more power resources are allocated by the BS. The methods that jointly optimize the BS transmit beamforming and phase shifts achieve about 10dB gain compared to the method without RIS. The three heuristic algorithms, including the simulated annealing algorithm, genetic algorithm and NGA, obtain much better performance than the sequantial search method. The performance loss of sequantial search is mainly due to the fact that decoupling the phase shifts and estimating them one by one is suboptimal. As expected, the NGA outperforms all comparison methods, indicating its effectiveness for hybrid beamforming design.

\begin{figure}[t]
	\centering
	\includegraphics[width=7cm,height=5.5cm]{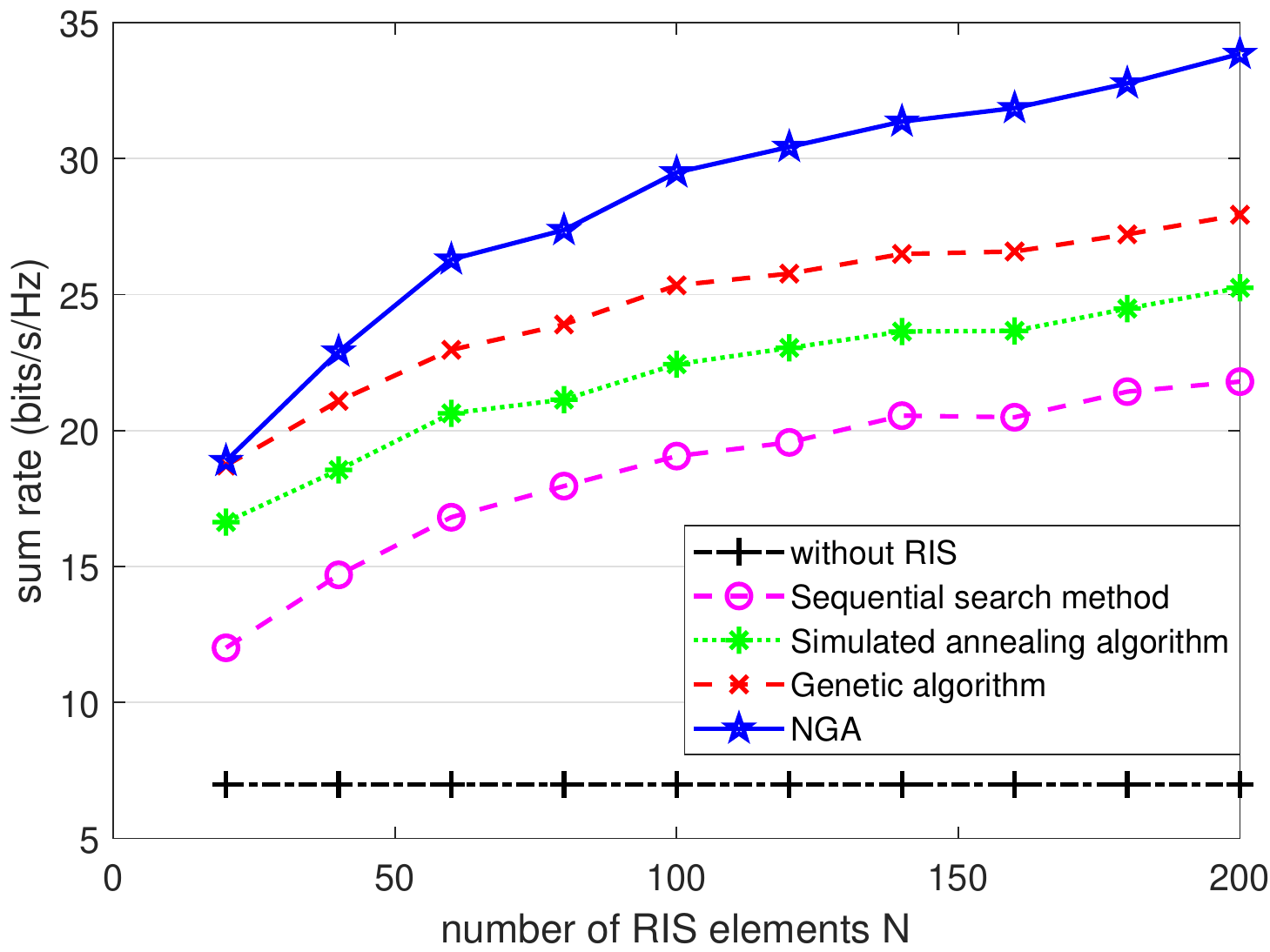}
	\caption{The sum rate versus the number of RIS elements $N$ when SNR=2dB and $b=2$.}
	\label{fig-reN}
\end{figure}

\textbf{Scalability over different numbers of RIS elements}. Fig. \ref{fig-reN} investigates the scalability of sum rate over the number of RIS elements $N$ with SNR=2dB and $b=2$. Except for the method without RIS, the remaining methods obtains higher sum rate as $N$ increases. Our NGA retains the highest sum rate with larger $N$ values. Particularly, as $N$ increases from 20 to 200, the proposed NGA achieves a surprising 15dB gain while the sequential search method, simulated annealing algorithm and genetic algorithm only have 9dB gain. The superiority performance of NGA is because the inherent niching technique owns an outstanding ability of exploring the search space, which avoids premature convergence and approaches the global optima as close as possible. We remark that the NGA is suggested to use in the cases with large-scale RISs ($N>20$).

\begin{figure}[t]
	\centering
	\includegraphics[width=7cm,height=5.5cm]{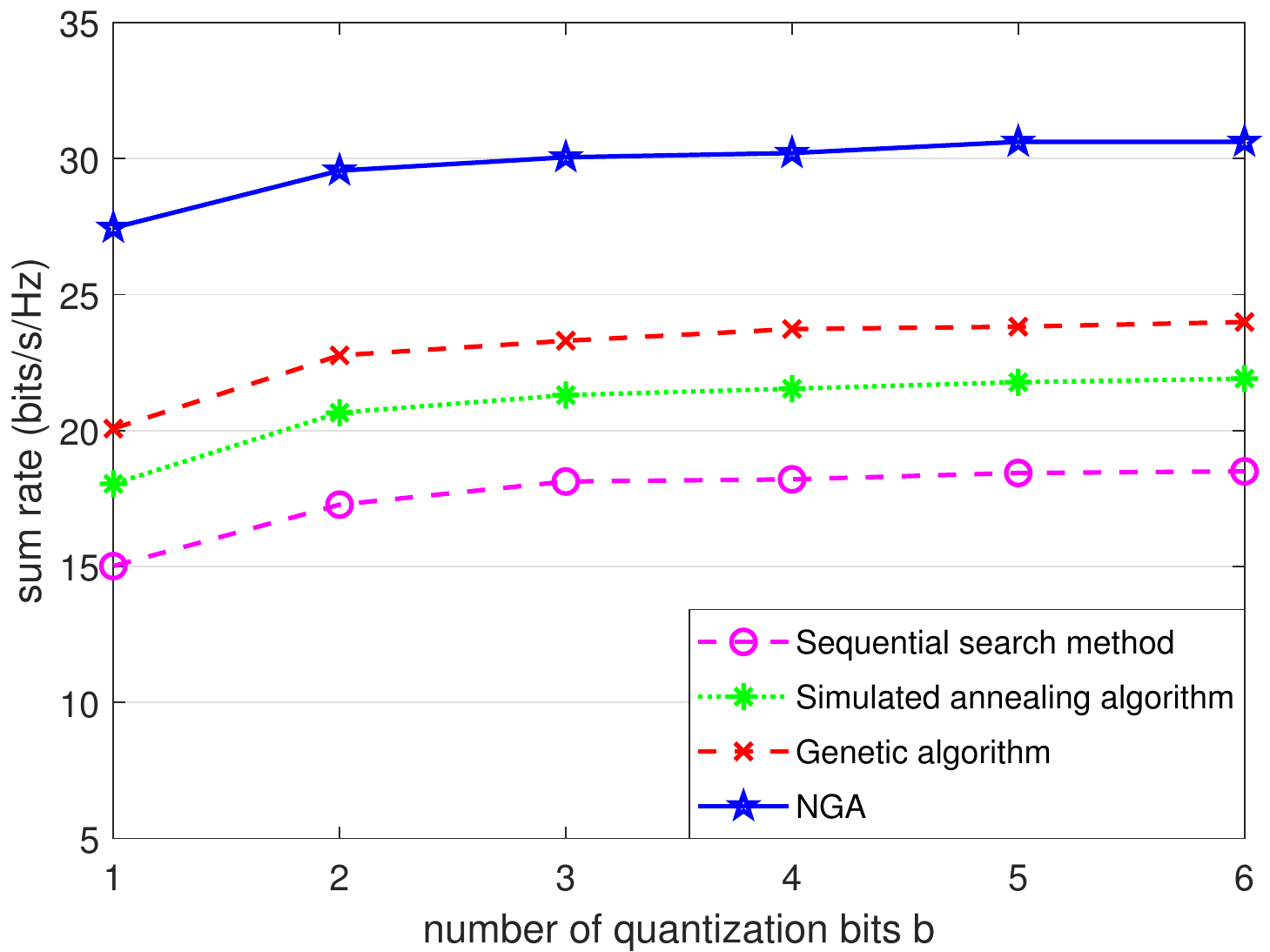}
	\caption{The sum rate versus the number of quantization bits $b$ when SNR=2dB and $N=100$.}
	\label{fig-reb}
\end{figure}

\textbf{Effect of different quantization bits}. Fig. \ref{fig-reb} depicts the sum rate of all users versus the number of quantization bits $b$ when SNR=2dB and $N=100$. Our proposed algorithm retains the highest sum rate under all numbers of quantization bits. As the number of quantization bits $b$ increases from 1 to 2, the sum rate of all algorithms grow fast, then basically remain unchanged when $b$ continues growing. It reveals that the system performance tends to be saturated when the $b$ exceeds 2. Therefore, 2-bit phase shifts is practically sufficient to achieve satisfying performance.

\section{Conclusion}
In this paper, we investigate the fitness landscape of the sum rate maximization problem in a RIS-aided downlink MU-MISO communication system. To explore the local and global properties of the fitness landscapes, a fitness distance correlation analysis and an autocorrelation analysis are performed. The analysis results exhibit a multi-modal, rugged landscape as the number of RIS elements increases or more quantization bits for the RIS are used. This explains why existing works easily get trapped into local optima or even fail in the case with large-scale RISs. 

Therefore, to handle the multi-modal landscape structure, we propose a novel niching genetic algorithm to jointly estimate the transmit beamforming at the BS and the discrete phase shifts at the RIS. Specially, the nearest-better clustering technique is incorporated to local multiple local optima and better find the global optima. A minimum species size is presented to enhance the convergence speed without influencing the performance.

Extensive simulation results fully demonstrate that the proposed method achieve significant capacity gains compared to existing works, particularly in the case with large-scale RISs.

\section*{Acknowledgment}

\ifCLASSOPTIONcaptionsoff
  \newpage
\fi


\bibliographystyle{IEEEtran}
\bibliography{Ref}

\end{document}